\documentstyle[]{article}

\newcounter{sentencectr}
\newcounter{sentencesubctr}

\renewcommand{\thesentencectr}{(\smainform{sentencectr})}
\renewcommand{\thesentencesubctr}{\thesentencectr\ssubform{sentencesubctr}}

\newcommand{\smainform}{\arabic}
\newcommand{\ssubform}{\alph}
\newcommand{\ssubpunc}{.{}}

\newcommand{\beginsentences}{ %
\pagebreak[3] %
\begin{list}{(\thesentencectr)}
   {\usecounter{sentencesubctr}
    \setlength{\topsep}{1ex}
    \setlength{\itemsep}{0 in}
    \setlength{\labelwidth}{0.5 in}
    \addtolength{\leftmargin}{4ex}
    \setlength{\labelsep}{.05in}
    \setlength{\parsep}{0 in}}}
\def\endsentences{\end{list}}

\newcommand{\sitem}{\renewcommand{\thesentencesubctr}{(\smainform{sentencectr}}
                    \refstepcounter{sentencectr}
     \item[(\smainform{sentencectr})\hfill]}
\newcommand{\smainitem}{\renewcommand{\thesentencesubctr
                                    }{\thesentencectr\ssubform{sentencesubctr}}
                        \setcounter{sentencesubctr}{0}
                        \refstepcounter{sentencectr}
                        \refstepcounter{sentencesubctr}
     \item[\thesentencectr\hfill\ssubform{sentencesubctr}\ssubpunc]}
\newcommand{\ssubitem}{\refstepcounter{sentencesubctr}
     \item[\hfill\ssubform{sentencesubctr}\ssubpunc]}

\makeatletter
\newcommand{\smainlabel}[1]{{
\renewcommand{\@currentlabel}{\thesentencectr}\label{#1}}}

\newcommand{\ssublabel}[1]{{
\renewcommand{\@currentlabel}{\ssubform{sentencesubctr}}\label{#1}}}
\makeatother

\makeatother

\columnwidth \textwidth
\makeatletter

\newcommand{\spacedcitations}{
\def\@citex[##1]##2%
{\if@filesw\immediate\write\@auxout{\string\citation{##2}}\fi
  \def\@citea{}\@cite{\@for\@citeb:=##2\do
    {\@citea\def\@citea{, }\@ifundefined
       {b@\@citeb}{{\bf ?}\@warning
       {Citation `\@citeb' on page \thepage \space undefined}}%
\hbox{\csname b@\@citeb\endcsname}}}{##1}}
}

\newcommand{\nobiblabels}{\def\@biblabel##1{}
 \@ifundefined{chapter}{\def\thebibliography##1{\section*{References\markboth
  {REFERENCES}{REFERENCES}}\list
  {[\arabic{enumi}]}{\setlength{\labelwidth}{1.5em}
    \setlength{\labelsep}{0em}
    \leftmargin\labelwidth
    \advance\leftmargin\labelsep
    \setlength{\itemindent}{-1.5em}
    \setlength{\itemsep}{0em}
    \setlength{\parsep}{0em}
    \usecounter{enumi}}
    \def\newblock{\hskip .11em plus .33em minus -.07em}
    \sloppy\clubpenalty4000\widowpenalty4000
    \sfcode`\.=1000\relax}}%
{\def\thebibliography##1{\chapter*{Bibliography\markboth
  {BIBLIOGRAPHY}{BIBLIOGRAPHY}}\list
  {[\arabic{enumi}]}{\setlength{\labelwidth}{1.5em}
    \setlength{\labelsep}{0em}
    \leftmargin\labelwidth
    \advance\leftmargin\labelsep
    \setlength{\itemindent}{-1.5em}
    \usecounter{enumi}}
    \def\newblock{\hskip .11em plus .33em minus -.07em}
    \sloppy\clubpenalty4000\widowpenalty4000
    \sfcode`\.=1000\relax}}}

\makeatother

\nobiblabels

\newcommand{\xbar}[1]{\mbox{#1$'$}}

\newtheorem{definition}{Definition}

\begin{document}

\vspace*{-1in}
\begin{center}
{\small
Appeared in {\em 3e Colloque International sur les grammaires
  d'Arbres Adjoints (TAG+3).}  Technical Report TALANA-RT-94-01,
  TALANA, Universit\'{e} Paris 7, 1994.}
\end{center}
\vspace*{.3in}

\begin{center}
{}From Regular to Context Free to Mildly Context Sensitive Tree
Rewriting Systems:\\
The Path of Child Language Acquisition\\

\vspace{8pt}

Robert Frank\\
University of Delaware\\
Department of Linguistics\\
46 E.~Delaware Avenue\\
Newark, DE 19716 USA\\
email: rfrank@cis.udel.edu
\end{center}

\begin{quote}
\begin{center}
Abstract
\end{center}
{\small Current syntactic theory limits the range of grammatical
variation so severely that the logical problem of grammar learning is
trivial. Yet, children exhibit characteristic stages in syntactic
development at least through their sixth year. Rather than positing
maturational delays, I suggest that acquisition difficulties are the
result of limitations in manipulating grammatical representations. I
argue that the genesis of complex sentences reflects increasing
generative capacity in the systems generating structural descriptions:
conjoined clauses demand only a regular tree rewriting system;
sentential embedding uses a context-free tree substitution grammar;
modification requires TAG, a mildly context-sensitive system.  }
\end{quote}

\noindent
{\bf I.} Some current views of natural language syntax localize all
cross-linguistic variation in a small set of finite-valued parameters.
This has the effect of eliminating certain in-principle learnability
problems that plagued earlier views of grammar.  Since the range of
possible hypotheses is restricted, the child will show an inductive
bias, that is, she will at times be driven to conclusions seemingly
stronger than those warranted by the input data.  Further,
learnability in the limit is guaranteed, since the number of possible
grammars is finite.  Finally, the limited amount and simplicity of
information to be acquired means that grammars should be acquired
quickly and easily.  Unfortunately, this beautiful picture conflicts
with what we know from empirical studies of language acquisition:
children exhibit characteristic developmental stages in their
acquisition of grammar.  Consequently, if such a parametric view of
syntactic variation is correct, children must be held back in their
attempts at syntactic learning by something other than the inherent
difficulty of the learning task.

I suggest that children's acquisitional difficulties result not from
problems of grammatical acquisition per se, but rather from their
limited abilities in manipulating grammatical representations.  In
particular, I argue that the sequence of certain stages in syntactic
development can best be understood as a reflection of ever increasing
generative complexity of the underlying formal grammatical systems
used by the child to construct her tree structure representations.

\vspace{4pt}
\noindent
{\bf II.} It has been widely reported in empirical studies of language
acquisition that different types of complex (i.e.~multi-clausal)
sentences vary with respect to the point at which children first
exhibit mastery of them.  Looking at the naturalistic production data
of four children, Bloom et al.~(1980)\nocite{Bloometal80} report that
the productive use of complex sentences involving conjunction
consistently precedes that of sentences involving complementation
which in turn precedes sentences involving relativization.  This
result is supported by experimental study of children's comprehension.
Tavakolian (1981)\nocite{Tavakolian81} demonstrates that young
children exhibit difficulty in interpreting relative clauses.  She
argues that the interpretations that children do assign to such
structures result from their (incorrectly) imposing a conjoined clause
analysis.  I take this tendency to prefer conjunction over
relativization to be the same effect observed by Bloom and her
colleagues.  Further, Goodluck (1981), Hsu et al. (1985), McDaniel and
Cairns (1990) among
others\nocite{McDaniel&Cairns90}\nocite{Goodluck81}\nocite{HCF85} have
found that children correctly interpret ``control constructions'' in
which the empty subject is within a complement clause, such
as~(1), at an earlier age than cases where the empty
subject is within an adverbial clause as in~(2).
\beginsentences
\smainitem Cookie Monster tells Grover$_i$ [PRO$_i$ to jump over the fence]

\ssubitem Grover$_i$ was told by Cookie Monster [PRO$_i$ to jump over
the fence]

\sitem Cookie Monster$_i$ touches Grover [after PRO$_i$ jumping over
the fence]
\endsentences
Let us suppose that a child's incorrect interpretations
in~(2) are the result of her inability to assign this
sentence a structural representation appropriate according to the
adult grammar.  If we assume that complex sentences containing
adverbial clause adjuncts are similar to sentences containing relative
clauses in the relevant structural respects, i.e., they involve
adjunction structures, this phenomenon can be seen as another instance
of Bloom et al.'s sequence of complementation before relativization
(now modification).

\vspace{4pt}
\noindent
{\bf III.} Thus far, we've seen that children's acquisition of complex
constructions proceeds according to the sequence coordination $<$
complementation $<$ modification.  Yet, we have not provided an
explanation for why these should be so ordered.  What I will now
suggest is that these stages are ordered by the ever greater demands
of generative capacity that they impose upon the formal tree rewriting
system used to construct phrase structure representations.

Before proceeding with this, we will require a brief detour into
defining a novel tree rewriting formalism.  Developing a suggestion of
Weir (1987)\nocite{Weir87} (though in a restricted fashion), let us
define a schematic tree grammar (STG) as a 4-tuple $G=(V_N, V_T, S,
I)$ where $V_N$ is a finite set of non-terminals, $V_T$ is a finite
set of terminals, $S$ is a distinguished non-terminal and $I$ is a
finite set of schematic initial trees.  The set of schematic initial
trees in an STG may be any finite set of finite tree structures whose
frontier nodes are drawn from $V_T \cup V_N$ and whose internal nodes
are drawn from $V_N$.  Further, all nodes of schematic initial trees
but the root may be annotated with the superscripts $+$ or $*$.  The
intuition behind the use of these superscripts is the same as their
usage in regular expressions: each schematic tree represents an
infinite set of trees, just as each regular expression represents an
infinite set of strings.  When a schematic tree contains a node $N$
marked by $+$, the class of trees representd by this schematic tree
includes trees containing 1 or more copies of the subtree dominated by
$N$, each copy attached to $N$'s parent.  Similarly, schematic trees
containing nodes marked $*$ correspond to those trees where this node
appears 0 or more times.  We formalize this as follows:
\begin{definition} A (possibly empty) sequence of trees $\langle
\tau_1, \ldots \tau_k\rangle$ {\em instantiates a schematic tree}
$\sigma$ iff:\\ 1. if the root of $\sigma$ is superscripted by $+$,
the sequence of $\tau_i$'s is of length $\ge 1$;\\ 2. if the root of
$\sigma$ is not superscripted, the sequence of $\tau_i$'s is of length
exactly 1;\\ 3. for each $\tau_i$, the root is labelled identically to
the root of $\sigma$;\\ 4. for each $\tau_i$, the sequence of subtrees
dominated by the children of the root of $\tau_i$, $\langle \tau_i^1,
\ldots \tau_i^n\rangle$, may be partitioned into a sequence of
contiguous subsequences, $\langle \tau_i^1, \ldots
\tau_i^j\rangle,\langle\tau_i^{j+1},\ldots\tau_i^{k}\rangle\ldots
\langle \tau_i^m,\ldots\tau_i^n\rangle$, so that these subsequences
successively instantiate the subtrees dominated by the children of
root of $\sigma$ from left to right.  \end{definition} Derivations in
an STG $G$ do not directly utilize the schematic trees in $I$, but
rather manipulate the trees which instantiate the trees in $I$.  The
only combinatory operation we allow in an STG is substitution.
Application of substitution is however restricted so as to prevent the
generation of recursive structures.  The set of derivable trees from a
grammar $G$, $D(G)$ is thus defined as follows: \begin{definition}
$\tau$ is {\em derivable} by $G$, $\tau \in D(G)$, iff:\\ 1. $\tau$
instantiates some $\sigma \in I$; or\\ 2. $\tau$ is the result of
substituting $\tau'$ into $\tau''$ where $\tau', \tau'' \in D(G)$ and
no non-terminal on the path from the root of $\tau''$ to the site of
subsitution appears in $\tau'$.  \end{definition} We finally define
the set of trees and strings which are generated by an STG $G$ as
follows: \begin{definition} The {\em tree set} of an STG $G$ is the
set of trees $T(G) = \{\tau | \tau \in D(G)$, $\tau$ is rooted in $S$
and the frontier of $\tau \in V_T^*\}$.  \end{definition}
\begin{definition} The {\em string language} of an STG $G$ is $L(G) =
\{ w | w$ is the frontier of some $\tau \in T(G) \} $ \end{definition}

Two things concerning this tree rewriting system are relevant for our
concerns.  The first is that the string languages generated by STGs
are all and only the regular languages. The second is that, in spite
of its relatively weak formal power, it is nonetheless sufficiently
expressive to express analyses of (certain cases of) coordination
structures.  If the schematic tree below on the left is in the
grammar, the STG will generate the string {\em John has eaten an apple
and Fred has eaten peaches and a candy bar}: the tree below on the
right, which duplicates the IP root node once and duplicates the NP
object, instantiates this schematic tree.  We need only perform the
relevant substitutions into the NP nodes to complete the
derivation.\footnote{In the interest of space, we put aside certain
complications concerning the insertion of conjunctions, the allowance
for some degree of lexical variation within the internal structure of
coordinated phrases.  Such issues can be dealt with without increase
in generative capacity.}

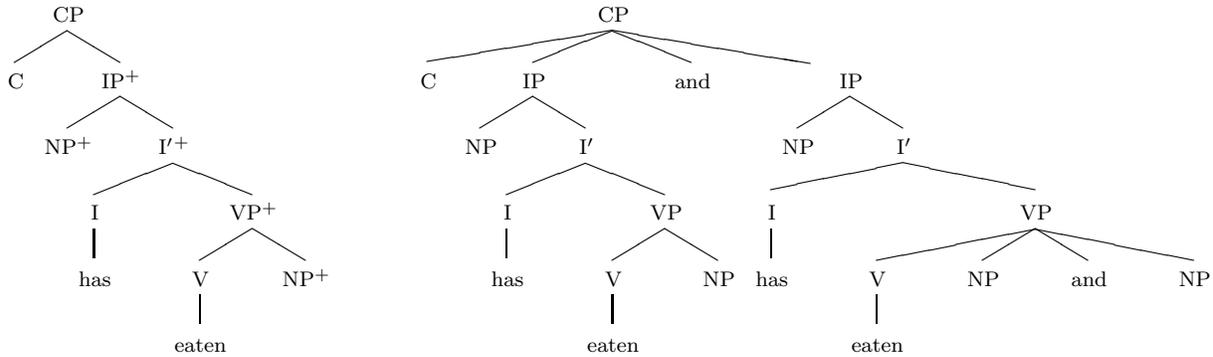
\mbox{\hspace{-.75in}
  \begin{picture}(200.0,149.0)(0,0)
   \put(75.0,136.0){\makebox(1,1){\rule[-1ex]{0ex}{3ex}{\footnotesize CP}}}
   \put(75.0,131.0){\line(-5,-3){20.0}}
   \put(75.0,131.0){\line(5,-3){20.0}}
   \put(55.0,111.0){\makebox(1,1){\rule[-1ex]{0ex}{3ex}{\footnotesize C}}}
   \put(95.0,111.0){\makebox(1,1){\rule[-1ex]{0ex}{3ex}{\footnotesize IP$^+$}}}
   \put(95.0,106.0){\line(-5,-3){20.0}}
   \put(95.0,106.0){\line(5,-3){20.0}}
   \put(75.0,86.0){\makebox(1,1){\rule[-1ex]{0ex}{3ex}{\footnotesize NP$^+$}}}
   \put(115.0,86.0){\makebox(1,1){\rule[-1ex]{0ex}{3ex}{\footnotesize
\xbar{I}$^+$}}}
   \put(115.0,81.0){\line(-5,-2){30.0}}
   \put(115.0,81.0){\line(5,-2){30.0}}
   \put(85.0,61.0){\makebox(1,1){\rule[-1ex]{0ex}{3ex}{\footnotesize I}}}
   \put(85.0,56.0){\line(0,-1){11.0}}
   \put(85.0,36.0){\makebox(1,1){\rule[-1ex]{0ex}{3ex}{\footnotesize has}}}
   \put(145.0,61.0){\makebox(1,1){\rule[-1ex]{0ex}{3ex}{\footnotesize VP$^+$}}}
   \put(145.0,56.0){\line(-5,-3){20.0}}
   \put(145.0,56.0){\line(5,-3){20.0}}
   \put(125.0,36.0){\makebox(1,1){\rule[-1ex]{0ex}{3ex}{\footnotesize V}}}
   \put(125.0,31.0){\line(0,-1){11.0}}
   \put(125.0,11.0){\makebox(1,1){\rule[-1ex]{0ex}{3ex}{\footnotesize eaten}}}
   \put(165.0,36.0){\makebox(1,1){\rule[-1ex]{0ex}{3ex}{\footnotesize NP$^+$}}}
   \put(0.0,0.0){\makebox(1,1)[l]{\rule[-1ex]{0ex}{3ex}{\normalsize}}}
  \end{picture}
\hspace{-.7in}
  \begin{picture}(350.0,149.0)(0,0)
   \put(125.0,136.0){\makebox(1,1){\rule[-1ex]{0ex}{3ex}{\footnotesize CP}}}
   \put(125.0,131.0){\line(-6,-1){70.0}}
   \put(125.0,131.0){\line(-5,-2){30.0}}
   \put(125.0,131.0){\line(5,-2){30.0}}
   \put(125.0,131.0){\line(6,-1){75.0}}
   \put(55.0,111.0){\makebox(1,1){\rule[-1ex]{0ex}{3ex}{\footnotesize C}}}
   \put(95.0,111.0){\makebox(1,1){\rule[-1ex]{0ex}{3ex}{\footnotesize IP}}}
   \put(95.0,106.0){\line(-5,-3){20.0}}
   \put(95.0,106.0){\line(5,-3){20.0}}
   \put(75.0,86.0){\makebox(1,1){\rule[-1ex]{0ex}{3ex}{\footnotesize NP}}}
   \put(115.0,86.0){\makebox(1,1){\rule[-1ex]{0ex}{3ex}{\footnotesize
\xbar{I}}}}
   \put(115.0,81.0){\line(-5,-2){30.0}}
   \put(115.0,81.0){\line(5,-2){30.0}}
   \put(85.0,61.0){\makebox(1,1){\rule[-1ex]{0ex}{3ex}{\footnotesize I}}}
   \put(85.0,56.0){\line(0,-1){11.0}}
   \put(85.0,36.0){\makebox(1,1){\rule[-1ex]{0ex}{3ex}{\footnotesize has}}}
   \put(145.0,61.0){\makebox(1,1){\rule[-1ex]{0ex}{3ex}{\footnotesize VP}}}
   \put(145.0,56.0){\line(-5,-3){20.0}}
   \put(145.0,56.0){\line(5,-3){20.0}}
   \put(125.0,36.0){\makebox(1,1){\rule[-1ex]{0ex}{3ex}{\footnotesize V}}}
   \put(125.0,31.0){\line(0,-1){11.0}}
   \put(125.0,11.0){\makebox(1,1){\rule[-1ex]{0ex}{3ex}{\footnotesize eaten}}}
   \put(165.0,36.0){\makebox(1,1){\rule[-1ex]{0ex}{3ex}{\footnotesize NP}}}
   \put(155.0,111.0){\makebox(1,1){\rule[-1ex]{0ex}{3ex}{\footnotesize and}}}
   \put(215.0,111.0){\makebox(1,1){\rule[-1ex]{0ex}{3ex}{\footnotesize IP}}}
   \put(215.0,106.0){\line(-5,-3){20.0}}
   \put(215.0,106.0){\line(5,-3){20.0}}
   \put(195.0,86.0){\makebox(1,1){\rule[-1ex]{0ex}{3ex}{\footnotesize NP}}}
   \put(235.0,86.0){\makebox(1,1){\rule[-1ex]{0ex}{3ex}{\footnotesize
\xbar{I}}}}
   \put(235.0,81.0){\line(-5,-1){50.0}}
   \put(235.0,81.0){\line(5,-1){50.0}}
   \put(185.0,61.0){\makebox(1,1){\rule[-1ex]{0ex}{3ex}{\footnotesize I}}}
   \put(185.0,56.0){\line(0,-1){11.0}}
   \put(185.0,36.0){\makebox(1,1){\rule[-1ex]{0ex}{3ex}{\footnotesize has}}}
   \put(285.0,61.0){\makebox(1,1){\rule[-1ex]{0ex}{3ex}{\footnotesize VP}}}
   \put(285.0,56.0){\line(-5,-1){60.0}}
   \put(285.0,56.0){\line(-5,-3){20.0}}
   \put(285.0,56.0){\line(5,-3){20.0}}
   \put(285.0,56.0){\line(5,-1){60.0}}
   \put(225.0,36.0){\makebox(1,1){\rule[-1ex]{0ex}{3ex}{\footnotesize V}}}
   \put(225.0,31.0){\line(0,-1){11.0}}
   \put(225.0,11.0){\makebox(1,1){\rule[-1ex]{0ex}{3ex}{\footnotesize eaten}}}
   \put(265.0,36.0){\makebox(1,1){\rule[-1ex]{0ex}{3ex}{\footnotesize NP}}}
   \put(305.0,36.0){\makebox(1,1){\rule[-1ex]{0ex}{3ex}{\footnotesize and}}}
   \put(345.0,36.0){\makebox(1,1){\rule[-1ex]{0ex}{3ex}{\footnotesize NP}}}
   \put(0.0,0.0){\makebox(1,1)[l]{\rule[-1ex]{0ex}{3ex}{\normalsize}}}
  \end{picture}
}

\noindent This regular tree rewriting system does not allow for the
generation of syntactic structures appropriate for sentential
complementation.  There is simply no way to produce the unbounded
degree of embedding that is necessary from a linguistic standpoint.
If, however, we move to the system of tree rewriting discussed by
Schabes (1990)\nocite{Schabes90} called Tree Substitution Grammars
(TSG), complementation can be accommodated.  A TSG consists of a
finite set of (finite) elementary trees whose leaves may be either
terminals or non-terminals.  As before, derivations consist in
substituting these elementary trees rooted in some category $C$ into
non-terminals at nodes along the frontiers of other elementary trees
also labeled $C$, but no restriction is imposed on possible recursion.
Schabes observes that TSGs are strictly context-free in their weak
generative capacity, though the tree sets they produce are somewhat
richer.\footnote{We point out that the addition of the device of
schematic trees to TSG or to TAG does not increase the weak generative
capacity since such node expansion can be simulated using substitution
or adjoining, though strong generative capaciy is affected since tree
structures of arbitrary arity cannot be generated without the use of
such schemas.  We leave open the issue of whether children's
grammatical systems (as well as those of adults) include such schema
all the way through development.}

Suppose now that we wish to generate sentences containing modification
structures, e.g.~relativization and adverbial adjuncts.  It is indeed
possible to use TSG to produce linguistically natural derived
structural representations.  We may however have independent
motivations for what may constitute an elementary tree in our grammar.
In particular, we might propose (following, among others, Frank
1992\nocite{Frank92}) that the elementary trees of a tree rewriting
system should contain only information concerning a single predicate,
such as a verb, and its associated argument structure.  If this is
true, then there can be no representation of a modification structure
(such as an adverbial modifier) in an elementary tree since it does
not play a role in the argument structure of the predicate heading
that tree.  Consequently, TSG will not be sufficient for our purposes.
Instead, we must turn to a somewhat more powerful system of tree
rewriting which allows the operation of adjoining, namely Tree
Adjoining Grammar (TAG -- Joshi, Levy and Takahashi
1975\nocite{JLT75}).  Adjoining allows us to introduce modification
structures into elementary trees which previously lacked any
representation of them.  Thus, TAG provides us with a richer and more
linguistically appealing class of derivations for a set of trees that
are generable by TSGs.  Note that the weak generative capacity of TAG
is strictly greater than the context-free power of TSG, though it is
nonetheless restricted to the class of so-called ``mildly
context-sensitive languages'' (Joshi, Vijay-Shanker and Weir
1991\nocite{Joshietal91}).

\vspace{4pt} \noindent {\bf IV.} To summarize, we have seen that there
is an increase in generative complexity associated with the tree
rewriting systems necessary for coordination, complementation and
modification.  I suggest that it is precisely this increase in
complexity that gives rise to this acquisitional sequence.  It is
important to observe that we are crucially dealing with complexity
measures in terms of tree rewriting systems here rather than the more
traditional string rewriting systems.  In building syntactic
representations, the core problem is the recovery and appropriate
factorization of dependencies.  These problems are most naturally
addressed, I would argue in a tree rewriting framework.  Further, from
the perspective of string language complexity, both coordination and
(right branching) complementation produce regular sets.  Consequently,
on the basis of a string complexity measure, we would not (contrary to
fact) predict any difference in acquisitional difficulty between these
cases. In previous work (Frank 1992), I found that other complex
sentences structures which are tied to the use of the adjoining
operation, but which are at least a priori distinct from modification,
show similar delayed acquisition.  This provides, I claim, independent
confirmation that our tree rewriting based approach is on the right
track.

Finally, we can ask whether this limited ability to manipulate systems
of tree rewriting is tied to computational load, following the
proposals of Joshi (1990)\nocite{Joshi90} and Rambow
(1992)\nocite{Rambow92}.  Certain experimental evidence suggests that
such an approach is correct: children's grammatical difficulties can
be alleviated to some degree if extraneous task demands are diminished
(cf.~Crain and Fodor 1993 for a review)\nocite{Crain&Fodor93}.  Thus,
if more powerful tree rewriting mechanisms are unavailable simply
because they demand too much of the child's resources, they might
become available when other resources are freed.


\begin{thebibliography}{}

\bibitem[\protect\citename{Bloom \bgroup et al.\ \egroup  }1980]{Bloometal80}
Bloom, Lois, Margaret Lahey, Lois Hood, Karin Lifter, and Kathleen Fiess.
\newblock 1980.
\newblock Complex sentences: Acquisition of syntactic connectives and the
  semantic relations they encode.
\newblock {\em Journal of Child Language}\/ 7:235--261.

\bibitem[\protect\citename{Crain and Fodor }1993]{Crain&Fodor93}
Crain, Stephen and Janet~D. Fodor. 1993.
\newblock Competence and performance in child language.
\newblock In E. Dromi (Ed.), {\em Language and cognition: A developmental
  perspective}\/. Ablex.

\bibitem[\protect\citename{Frank }1992]{Frank92}
Frank, Robert. 1992.
\newblock {\em Syntactic Locality and Tree Adjoining Grammar: Grammatical,
  Acquisition and Processing Perspectives}\/.
\newblock PhD thesis, University of Pennsylvania.

\bibitem[\protect\citename{Goodluck }1981]{Goodluck81}
Goodluck, Helen. 1981.
\newblock Children's grammar of complement-subject interpretation.
\newblock In S. Tavakolian (Ed.), {\em Language Acquisition and Linguistic
  Theory}\/. MIT Press.

\bibitem[\protect\citename{Hsu \bgroup et al.\ \egroup  }1985]{HCF85}
Hsu, Jennifer~Ryan, Helen~Smith Cairns, and Robert Fiengo.
\newblock 1985.
\newblock The development of grammars underlying children's interpretation of
  complex sentences.
\newblock {\em Cognition}\/ 20.

\bibitem[\protect\citename{Joshi }1990]{Joshi90}
Joshi, Aravind~K.
\newblock 1990.
\newblock Processing crossed and nested dependencies: an automaton perspective
  on the psycholinguistic results.
\newblock {\em Language and Cognitive Processes}\/ 5.

\bibitem[\protect\citename{Joshi \bgroup et al.\ \egroup  }1975]{JLT75}
Joshi, Aravind~K., Leon Levy, and Masako Takahashi.
\newblock 1975.
\newblock Tree adjunct grammars.
\newblock {\em Journal of the Computer and System Sciences}\/ 10.

\bibitem[\protect\citename{Joshi \bgroup et al.\ \egroup  }1991]{Joshietal91}
Joshi, Aravind~K., K. Vijay-Shanker, and David Weir. 1991.
\newblock The convergence of mildly context-sensitive grammatical formalisms.
\newblock In P. Sells, S. Shieber, and T. Wasow (Eds.), {\em Foundational
  Issues in Natural Language Processing}\/. Cambridge, MA: MIT Press.

\bibitem[\protect\citename{McDaniel and Cairns }1990]{McDaniel&Cairns90}
McDaniel, Dana and Helen~Smith Cairns. 1990.
\newblock Processing and acquisition of control structures by young children.
\newblock In L. Frazier and J. {de Villiers} (Eds.), {\em Language Processing
  and Language Acquisition}\/. Kluwer.

\bibitem[\protect\citename{Rambow }1992]{Rambow92}
Rambow, Owen. 1992.
\newblock A linguistic and computational analysis of the third construction.
\newblock In {\em Proceedings of the 30th Annual Meeting of the Association for
  Computational Linguistics}\/, Newark, DE.

\bibitem[\protect\citename{Schabes }1990]{Schabes90}
Schabes, Yves. 1990.
\newblock {\em Mathematical and Computational Aspects of Lexicalized
  Grammars}\/.
\newblock PhD thesis, University of Pennsylvania.

\bibitem[\protect\citename{Tavakolian }1981]{Tavakolian81}
Tavakolian, Susan. 1981.
\newblock The conjoined clause analysis of relative clauses.
\newblock In S. Tavakolian (Ed.), {\em Language Acquisition and Linguistic
  Theory}\/. MIT Press.

\bibitem[\protect\citename{Weir }1987]{Weir87}
Weir, David. 1987.
\newblock From context-free grammars to tree adjoining grammars and beyond.
\newblock Technical Report MS-CIS-87-42 (LINC Lab 65), Departent of Computer
  and Information Sciences, University of Pennsyvania.
\newblock Dissertation Proposal.

\end{thebibliography}
\end{document}